\documentclass[manuscript,screen,acmsmall]{acmart}

\AtBeginDocument{%
  }

\usepackage{color}
\usepackage{listings}
\usepackage[ruled, vlined, linesnumbered]{algorithm2e}
\usepackage{wrapfig}
\usepackage{booktabs}  % 提供 \Xhline
\usepackage{graphicx}  % 提供 \resizebox
\usepackage{subcaption}

\definecolor{dmlgreen}    {RGB}{51,  160,  44}
\definecolor{dmlblue}     {RGB}{31,  120, 180}
\definecolor{dmlred}      {RGB}{202,   0,  32}
\definecolor{brown}       {RGB}{139,  69,  19}

\definecolor{deepblue}{rgb}{0,0,0.5}
\definecolor{deepred}{rgb}{0.6,0,0}
\definecolor{deepgreen}{rgb}{0,0.5,0}
\definecolor{mauve}{rgb}{0.58,0,0.82}
\definecolor{light-gray}{gray}{0.96}
\definecolor{aliceblue}{rgb}{0.94, 0.97, 1.0}
\usepackage[frozencache,cachedir=minted-cache]{minted}
\newcommand{\lstbg}[3][0pt]{{\fboxsep#1\colorbox{#2}{\strut #3}}}
\lstdefinelanguage{diff}{
  backgroundcolor=\color{aliceblue},           
  emph={},          
  emphstyle=\small\color{dmlred},  
  belowcaptionskip=0.7\baselineskip,
  aboveskip=0mm,
  belowskip=3mm,
  showstringspaces=false,
  columns=flexible,
  basicstyle={\linespread{1.1}\fontencoding{T1}\scriptsize\fontfamily{lmtt}\fontseries{m}\selectfont},
  numbers={left},
  xleftmargin={2em},%
  breaklines=true,
  breakatwhitespace=true,
  tabsize=3,
  morecomment=[f][\lstbg{red!20}]-,
  morecomment=[f][\lstbg{green!20}]+,
  morecomment=[f][\textit]{@@},
}

\lstnewenvironment{env_cpp}[2]
{
\lstset{
  language=C++,
  backgroundcolor=\color{gray!10}, %%%%%%%
  otherkeywords={ap_int},
  escapeinside={<@}{@>},
  escapechar=!,
  keywordstyle=\ttb\color{deepblue},%
  emph={},     
  emphstyle=\footnotesize\color{dmlred},  
  belowcaptionskip=0.7\baselineskip,
  aboveskip=-2mm,
  belowskip=5mm,
  showstringspaces=false,
  columns=flexible,
  basicstyle={\linespread{1.1}\fontencoding{T1}\footnotesize\fontfamily{lmtt}\fontseries{m}\selectfont},
  numbers={left},
  xleftmargin={2em},%
  framexleftmargin={1em},
  numberstyle=\tiny\color{gray},
  keywordstyle=\color{blue},
  commentstyle=\color{deepgreen},
  stringstyle=\color{mauve},
  breaklines=true,
  showspaces = false,
  showstringspaces = false,
  tabsize=3,
  frame=tb, 
  caption= #1, 
  label= #2
}%lstset
  \vspace{\baselineskip}
}%lstnewenvironment
{}
\newcommand\code[1]{\mintinline{python}{#1}}

\newcommand{\oursys}{{\textsc{TileLang}}\xspace}

\renewcommand\footnotetextcopyrightpermission[1]{}
\begin{document}

\title{\oursys{}: A Composable Tiled Programming Model for AI Systems}


% \author{tilelang team}

\author{Lei Wang$^{\mathsection}$}
% \thanks{*Part of this work was done during internships at Microsoft Research.}
\thanks{${\mathsection}$Equal contributions.}
\email{leiwang1999@outlook.com}
\affiliation{
  \institution{Peking University}
  \city{Beijing}
  \country{China}
}

\author{Yu Cheng$^{\mathsection}$}
\email{yucheng@pku.edu} % Replace with actual email
\affiliation{
  \institution{Peking University}
  \city{Beijing}
  \country{China}
}

\author{Yining Shi$^{\mathsection}$}
\email{yiningshi@pku.edu}
\affiliation{
  \institution{Peking University}
  \city{Beijing}
  \country{China}
}

\author{Zhengju Tang}
\email{zhengjutang@pku.edu} % Replace with actual email
\affiliation{
  \institution{Peking University}
  \city{Beijing}
  \country{China}
}

\author{Zhiwen Mo}
\email{zhiwen.mo25@imperial.ac.uk} % Replace with actual email
\affiliation{
  \institution{Imperial College London}
  \city{London}
  \country{United Kingdom}
}

\author{Wenhao Xie}
\email{wenhao@stu.pku.edu} % Replace with actual email
\affiliation{
  \institution{Peking University}
  \city{Beijing}
  \country{China}
}

\author{Lingxiao Ma}
\email{lingxiaoma@microsoft.com}
\affiliation{
  \institution{Microsoft Research}
  \city{Beijing}
  \country{China}
}

\author{Yuqing Xia}
\email{yuqingxia@microsoft.com} % Replace with actual email
\affiliation{
  \institution{Microsoft Research}
  \city{Beijing}
  \country{China}
}

\author{Jilong Xue}
\email{jilongxue@microsoft.com}
\affiliation{
  \institution{Microsoft Research}
  \city{Beijing}
  \country{China}
}

\author{Fan Yang}
\email{fanyang@microsoft.com}
\affiliation{
  \institution{Microsoft Research}
  \city{Beijing}
  \country{China}
}

\author{Zhi Yang}
\email{zhiyang@pku.edu} % Replace with actual email
\affiliation{
  \institution{Peking University}
  \city{Beijing}
  \country{China}
}

\renewcommand{\shortauthors}{Lei et al.}

\begin{abstract}

Modern AI workloads rely heavily on optimized computing kernels for both
training and inference. These AI kernels follow well-defined data-flow patterns,
such as moving tiles between DRAM and SRAM and performing a sequence of
computations on those tiles. However, writing high-performance kernels remains
complex despite the clarity of these patterns. Achieving peak performance requires
careful, hardware-centric optimizations to fully leverage modern accelerators. While
domain-specific compilers attempt to reduce the burden of writing high-performance
kernels, they often struggle with usability and expressiveness gaps.

In this paper, we present \oursys{}, a generalized tiled programming model for more efficient AI Kernel programming. \textbf{\oursys{} decouples scheduling space (thread binding, layout, tensorize and pipeline) from dataflow, and encapsulated them as a set of customization annotations and primitives.} This approach allows users to focus on the kernel’s data-flow itself, while leaving most other optimizations to compilers. We conduct comprehensive experiments on commonly-used devices, across numerous experiments, our evaluation shows that \oursys{} can achieve state-of-the-art performance in key kernels, demonstrating that its unified block-and-thread paradigm and transparent scheduling capabilities deliver both the power and flexibility demanded by modern AI system development.

\end{abstract}


\received{20 February 2007}
\received[revised]{12 March 2009}
\received[accepted]{5 June 2009}


\maketitle

\section{INTRODUCTION}

Over the past few years, the pursuit of higher performance in AI workloads\cite{openaichatgpt,GoogleBard2023,NewBing2023,yang2023diffusion} has accelerated the development of specialized kernels\cite{dao2022flashattention,nvidia2024cutlass,composable_kernel,thunderkittens} that drive both training and inference. Matrix multiplication, in particular, underpins a broad spectrum of neural network architectures, from straightforward feed-forward layers to massive Transformer-based models. To address the significant computational burden of these networks, custom kernels such as FlashAttention\cite{shah2024flashattention} have emerged to optimize attention mechanisms, reducing memory overhead and enhancing processing throughput. Nonetheless, achieving high efficiency on evolving accelerator hardware hinges on a nuanced blend of hardware-aware design and intricate tuning—challenges that have spurred a growing interest in more expressive domain-specific compilers.

Deep learning kernels are typically represented as data-flow patterns that
involve moving tiles between DRAM and SRAM and executing sequences of
computations on these tiles. Despite the apparent clarity of these patterns,
crafting high-performance kernels remains challenging because developers must
manually address several key optimizations:

\begin{itemize} 
    \item \textbf{Thread Binding.} Binding refers to the process of mapping
    tile operations and data to the appropriate thread. In modern accelerator architectures—such as GPUs—this involves the careful allocation of tasks across thread blocks, warps,
    and individual threads to maximize parallelism and minimize load imbalance. An optimal
    binding strategy enhances data locality and reduces overhead associated with thread
    synchronization and divergence, thereby contributing to improved computational
    throughput.

    \item \textbf{Memory Layout.} Memory layout optimization entails the systematic organization of data in physical memory to eliminate bank conflicts and ensure efficient access patterns. As demonstrated by recent work \cite{phothilimthana2019swizzle,hagedorn2023graphene}, this process often requires transforming the natural data representation into a tiled or blocked format that aligns with the architecture’s memory subsystem. Such reorganization facilitates coalesced accesses and effective cache utilization, thereby reducing memory latency and enhancing overall system performance.

    \item \textbf{Intrinsic Tensorization.} Leveraging intrinsic functions entails the direct utilization of target-specific instructions optimized for performance. Modern processors and accelerators provide specialized operations—such as Tensor Core\cite{tensorcore} and Matrix Core\cite{matrixcore}—that can perform multiple arithmetic operations simultaneously, along with mechanisms like vector copy and asynchronous copy to better utilize bandwidth. Employing these intrinsic instructions requires precise management of data types, memory alignment, and control flow to fully exploit the hardware’s computational capabilities, leading to significant speedups in critical kernel operations.
    
    \item \textbf{Pipeline.} Pipelining is the technique of overlapping data movement with computation to mitigate memory access latencies. By concurrently scheduling data transfers and computational tasks, pipelining ensures that processing units remain active and that idle periods due to memory latency are minimized. In advanced Nvidia Hopper architecture, Tensor Memory Accelerator (TMA)\cite{nvidia2023h100} can facilitate this process by enabling asynchronous process for different compute units—such as CUDA Cores and Tensor Cores—further enhancing concurrency. 
\end{itemize}

Although recent domain-specific compilers for AI workloads~\cite{chen2018tvm,ansor,roller} have greatly simplified the creation of high-performance kernels, they still intertwine most low-level optimizations with the kernel implementation, even when the dataflow is explicitly exposed. Triton \cite{triton}, for example, supplies intuitive block-level primitives but hides thread behavior, memory layout, and address-space annotations behind automatically generated strategies. This abstraction eases programming, yet it hampers experienced developers who seek to extract maximum performance—for instance, when implementing matrix multiplication with quantized weights. Such kernels typically demand inline assembly to perform vectorized datatype conversions \cite{kim2022whosays} and custom data layouts carefully aligned with hardware-specific memory buffers \cite{wang2024ladder}. While Triton provides vectorized operations such as \texttt{tl.dot}, extending them to bespoke use cases—e.g., by registering handcrafted high-performance tile operators through PTX—remains cumbersome. Furthermore, even though Triton exposes a user-friendly pipeline knob (\texttt{num\_stage}), it does not allow users to define an entirely custom pipeline. Consequently, domain experts are constrained in developing kernels that require explicit control over memory hierarchies and other fine-grained optimizations.

To address these limitations, we propose \oursys{}, a programming model that retains the simplicity of Triton while offering even greater flexibility. \oursys{} is designed to provide users with fine-grained control over the scheduling space to achieve higher performance. We argue that a key enabler for this is the decoupling of dataflow and scheduling: users focus solely on defining the dataflow using composable tile operators, while the compiler is responsible for exploring and applying scheduling strategies. When the compiler’s default optimizations fall short, users can exert more precise control at the frontend. We introduce a composable tiled programming abstraction in which core computation patterns—such as GEMM, COPY, ATOMIC, and REDUCE—are expressed using tile operators. These operators define the kernel’s dataflow independently of scheduling decisions. In parallel, a set of scheduling primitives and annotations are provided to capture further optimizations, giving users the option to either rely on compiler-generated schedules or manually fine-tune performance-critical aspects of the kernel.

To improve the usability of \oursys{}, we have implemented the frontend language in Python for a flexible programming style with minimal type annotations. Additionally, we introduce a compiler for \oursys{} that translates user-defined programs into highly optimized low-level code for efficient execution on modern hardware. The compiler automates key optimizations, reducing the manual effort required for performance tuning. In summary, our contributions are as follows:

\begin{enumerate}
    \item \textbf{Tile-Level Programming Language.} We designed a tile-level programming language that allows users to explicitly declare the placement of buffers within the hardware memory hierarchy. By leveraging a Layout Inference mechanism, the system abstracts away the complexity of efficiently parallelizing buffer operations while exposing thread-level control interfaces, enabling experts to precisely manage how each thread interacts with the buffers. 
    
    \item \textbf{Compiler with Automated Optimization.} We provided an accompanying compiler for \oursys{}, which includes a series of automated compilation passes. These passes encompass features such as automatic parallelization through a Layout Inference mechanism, dynamic parameter simplification for kernel libraries, automatic pipeline derivation, and loop tail splitting optimizations for dynamic shapes. This compiler ensures that \oursys{} programs are both highly efficient and easy to write. 
    
    \item \textbf{State-of-the-Art Performance.} Empirical evaluations on real-world AI kernels demonstrate that \oursys{} achieves performance comparable to, and sometimes exceeding, that of specialized vendor libraries and other DSL-based approaches such as Triton, across both NVIDIA and AMD GPUs.
\end{enumerate}

In the remainder of this paper, we present the design and implementation of TileLang. We begin by describing the language syntax and underlying programming model. We then detail the TileLang JIT compiler architecture, covering both hardware-agnostic and hardware-aware optimizations. Finally, we compare TileLang against existing efforts and conclude by summarizing our findings and outlining future directions for this unified approach to high-performance AI kernel development. We have open-sourced \oursys{}\footnote{\url{https://github.com/tile-ai/tilelang}}.

\section{A \oursys{} Example}

\begin{figure}[!htbp]
    \centering
    \includegraphics[width=\linewidth]{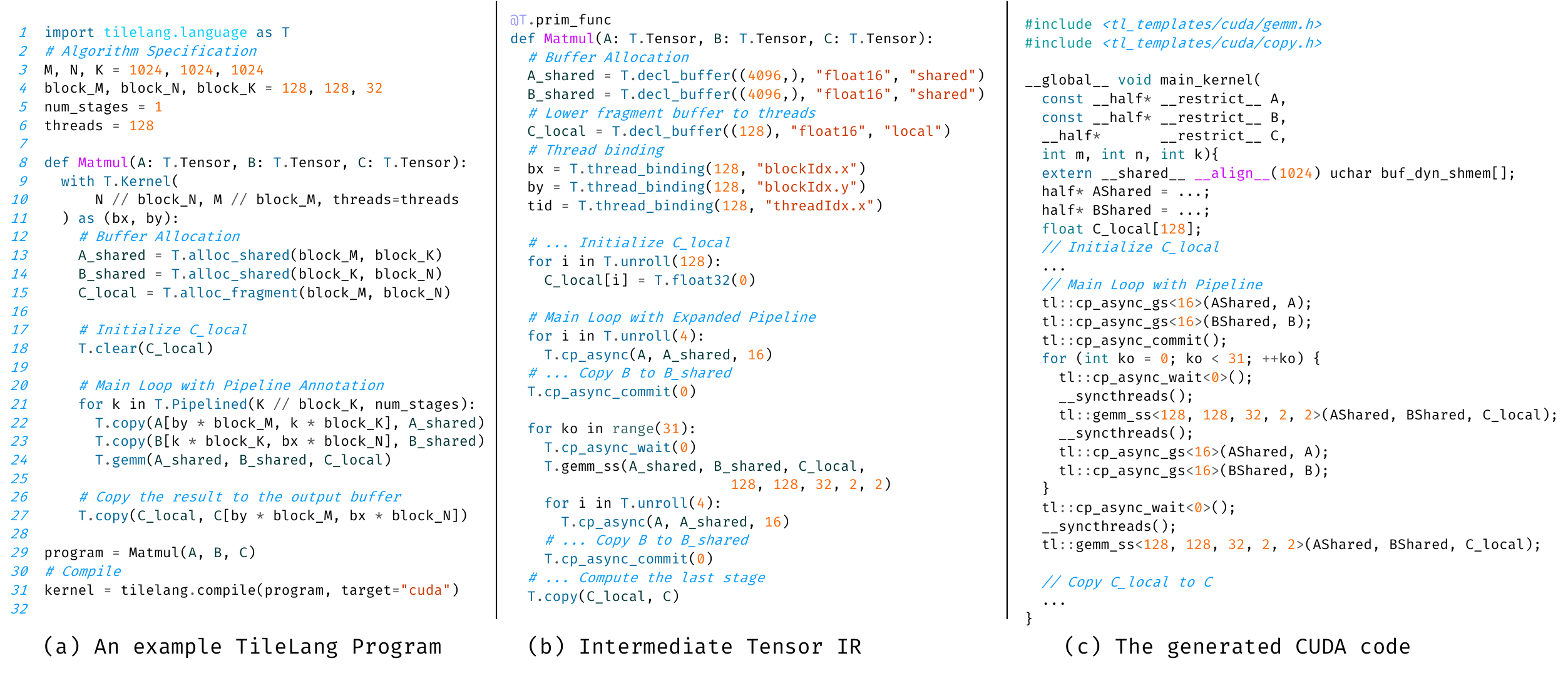}
    \caption{An example \oursys{} program and the corresponding lowered ir and generated cuda c code. The code snippets are simplified for demonstration purposes.}
    \label{fig:matmul_example}
\end{figure}

% \textbf{Examples:} \code{if-else}, \code{def <func>[<type params>](<args>)}, \code{T.alloc_fragments}, 

% \code{T.Pipelined}

Existing machine learning compilers that separate scheduling from computation, such as TVM, require users to explicitly distinguish between computation and scheduling. Additionally, users must manually register new tensor instructions and specify buffer layouts to achieve optimal performance. However, writing and understanding scheduling programs remains challenging. Although modern frameworks like Triton allow users to focus on tile-level programming, their dataflow representation is often unclear, and they require the use of certain workarounds—such as masked conditional loads—or hardware-specific features like Tensor Memory Accelerator (TMA). While frameworks such as ThunderKitten abstract programs into a tile-granular combination of load, compute, store, and synchronization operations, their dataflow remains insufficiently transparent, limiting users' ability to apply further optimizations. Moreover, with the widespread adoption of Python-based deep learning frameworks~\cite{pytorch, wolf2019huggingface}, manually translating models into C++ for optimization is impractical. Therefore, in designing \oursys{}, we emphasize three key principles: (1) \textbf{Pythonic design}, which integrates seamlessly with the Python ecosystem, providing a familiar coding experience and reducing the learning curve; (2) \textbf{Dataflow-centric}, which enables users to focus primarily on dataflow while abstracting away low-level scheduling complexities. It decouples scheduling aspects—such as thread binding, memory layout, tensorization, and pipelining—from dataflow, encapsulating them as a set of customizable annotations and primitives to enhance both programmability and maintainability; and (3) \textbf{Composability}, ensuring that kernels, primitives, and scheduling strategies can be seamlessly combined to construct complex designs.

In the following, we implement a general matrix multiplication (GEMM) kernel in \oursys{} to illustrate its basic syntax and demonstrate how it enhances productivity. As shown in Figure~\ref{fig:matmul_example}(a), the implementation begins by defining the GEMM kernel's inputs and outputs (Line 8), specifying their shapes and data types. Subsequently, we initialize the kernel context (Lines 9–11), which determines the grid size and total number of threads, followed by the kernel body (Lines 12–27), which includes on-chip memory allocations and data flow management. Since \oursys{} is a Python-embedded programming language, it supports all imperative constructs of Python (e.g., \code{if-else}, \code{for}, and \code{while}), with the key distinction that users must provide explicit type annotations for function arguments and variable declarations. This requirement arises due to Python’s dynamic typing, which may not be inherently suitable for device code generation (e.g., CUDA/HIP), where static data types are essential for determining precise data bitwidths. In \oursys{}, type annotations explicitly define element types and tensor shapes, ensuring correctness and efficient code generation. Additionally, \oursys{} allows explicit memory allocation, providing greater control over data placement and access patterns. In the given implementation, \oursys{} employs \code{T.alloc_shared} to store submatrices of \( A \) and \( B \) in shared memory, while \code{T.alloc_fragments} is used to allocate accumulators in register files at the block level. Furthermore, the use of pipelined execution (\code{T.Pipelined}) enables the overlapping of memory transfers with computation, effectively hiding memory latency and improving overall throughput. The \code{T.gemm} operation leverages NVIDIA CUTLASS or manually written HIP code to perform tile-level matrix computation efficiently. By automating low-level scheduling and synchronization, \oursys{} allows developers to focus on algorithm design rather than hardware-specific optimizations, thereby enhancing productivity while maintaining computational efficiency.  

Finally, we invoke \texttt{tilelang.compile} (Line 31) to lower the \texttt{tilelang} program into an intermediate representation (IR), as illustrated in Figure~\ref{fig:matmul_example}(b). This IR is then further compiled into an executable, generating the final optimized code, as shown in Figure~\ref{fig:matmul_example}(c).

\section{The Tile Language}

In this section, we introduce the foundations of our tile-based programming model, explain how \oursys{} systematically manages AI kernel development efficiently, and outline \oursys{}'s design philosophy of separating data flow from other scheduling spaces.

\begin{figure}[!htbp]
    \centering
    \includegraphics[width=\linewidth]{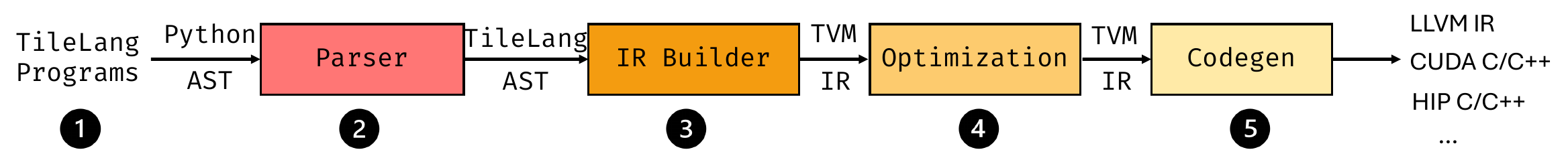}
    \vspace{-0.7cm}
    \caption{Stages of \oursys{} Compile Pipeline.}
    \vspace{-0.2cm}
    \label{fig:compile_pipeline}
\end{figure}

Figure~\ref{fig:compile_pipeline} illustrates the five-stage compilation pipeline of \oursys{}. Initially, developers write high-level programs using TileLang to describe computational logic and data access patterns. In the Parser stage, TileLang programs are parsed into Python AST and subsequently transformed into TileLang AST. Next, the IR Builder converts the AST into TVM intermediate representation (IR), enabling us to leverage TVM's syntax tree and related infrastructure. Following this, the Optimization stage performs a series of graph optimizations and scheduling transformations to enhance execution efficiency. Finally, the Codegen stage translates the optimized IR into backend code such as LLVM IR, CUDA C/C++, or HIP C/C++, supporting various hardware platforms.

Table~\ref{tab:operators_and_primitives} showcases a representative subset of the dataflow operators and scheduling primitives provided by \oursys{}. The Tile Language embraces a data-centric programming paradigm, where core computational semantics are expressed through tile-level operators such as \texttt{T.copy}, \texttt{T.gemm}, and \texttt{T.reduce}. Complementing these operators, \oursys{} exposes a set of scheduling primitives that allow developers to fine-tune performance-critical aspects such as parallelism, pipelining, and memory layout. We will explain the design of these two components in the following sections.

\begin{table}[h!]
\caption{A partial list of the dataflow operators and scheduling primitives supported by \oursys{}.}
\label{tab:operators_and_primitives}
\scriptsize
\begin{tabular}{p{0.08\linewidth}p{0.31\linewidth}p{0.14\linewidth}p{0.36\linewidth}}
\toprule
\multicolumn{2}{c}{\textbf{Dataflow Centric Tile Operators}} & \multicolumn{2}{c}{\textbf{Scheduling Primitives}} \\
\midrule
\texttt{T.copy} & A specialized memory copy operator that abstracts parallel data movement among registers, shared memory, and global memory. 
& \texttt{T.Parallel} & Automates parallelization of loop iterations, mapping them to hardware threads, can also enable vectorization for additional performance gains. \\
\midrule
\texttt{T.gemm} & Automatically selects implementations (cute/cuda/hip) for high-performance matrix multiplication on different GPUs. 
& \texttt{T.Pipelined} & Enables loop-level pipelining to overlap data transfers with computation and supports hardware-specific instructions such as async copy and TMA. \\
\midrule
\texttt{T.reduce} & A flexible reduction operator (e.g., sum, min, max) exploiting warp- and block-level parallelism. 
& \texttt{T.annotate\_layout} & Allows the definition of custom memory layouts to minimize bank conflicts and optimize thread binding. \\
\midrule
\texttt{T.atomic} & Provides atomic operations (e.g., add, min, max) to ensure thread-safe updates in shared or global memory. 
& \texttt{T.use\_swizzle} & Improves L2 cache locality via swizzle thread blocks. \\
\bottomrule
\end{tabular}
\end{table}

\subsection{Tile-based Programming Model}
\label{sec:tile_based_programming_model}

Figure~\ref{fig:matmul_example} provides a concise matrix multiplication (GEMM) example in \oursys{}, illustrating how developers can employ high-level constructs such as tiles, memory placement, pipelining, and operator calls to manae data movement and computation with fine-grained control. 
In particular, this snippet Figure~\ref{fig:matmul_example}(a) demonstrates how multi-level tiling leverages different memory hierarchies (global, shared, and registers) to optimize bandwidth utilization and reduce latency. Overall, Figure~\ref{fig:matmul_example} (b) showcases how the Python-like syntax of \oursys{} allows developers to reason about performance-critical optimizations within a user-friendly programming model.

\begin{figure}[!htbp]
    \centering
    \includegraphics[width=\linewidth]{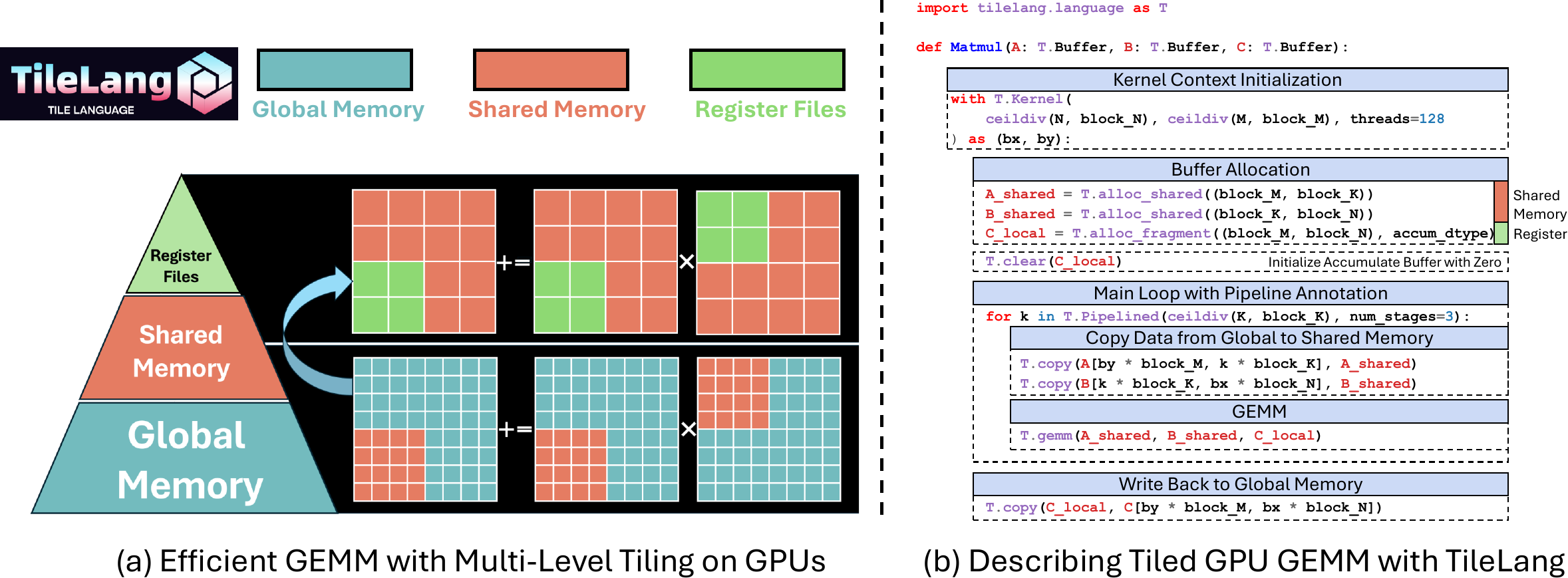}
    \vspace{-0.5cm}
    \caption{Optimizing GEMM with Multi-Level Tiling on GPUs via \oursys{}.}
    \vspace{-0.5cm}
    \label{fig:matmul_example}
\end{figure}

\paragraph{Tile declarations.} 
At the heart of our approach is the notion of \emph{tiles} as first-class objects in the programming model. A tile represents a shaped portion of data, which can be owned and manipulated by a warp, thread block, or equivalent parallel unit. In the \texttt{Matmul} example, the \texttt{A} and \texttt{B} buffers are read in tiled chunks (determined by \texttt{block\_M}, \texttt{block\_N}, \texttt{block\_K}) inside the kernel loop. With \texttt{T.Kernel}, \oursys{} defines the execution context, which includes the thread block index (\texttt{bx} and \texttt{by}) and the number of threads. These contexts can help us compute the index for each thread block, and making it easier for the \oursys{} to automatically inference and optimize memory access and computation. Additionally, these contexts allow users to manually control the behavior of each independent thread within a thread block.

\paragraph{Explicit Hardware Memory Allocation.}
A hallmark of \oursys{} is the ability to explicitly place these tile buffers in the hardware memory hierarchy. Rather than leaving it to a compiler's opaque optimization passes, \oursys{} exposes user-facing intrinsics that map directly to physical memory spaces or accelerator-specific constructs. In particular:
\begin{itemize}
\item \textbf{T.alloc\_shared}: Allocates memory in a fast, on-chip storage space, which corresponds to shared memory on NVIDIA GPUs. Shared memory is ideal for caching intermediate data during computations, as it is significantly faster than global memory and allows for efficient data sharing between threads in the same thread block. For example, in matrix multiplication, tiles of matrices can be loaded into shared memory to reduce global memory bandwidth demands and improve performance.
\item \textbf{T.alloc\_fragment}: Allocates accumulators in fragment memory, which corresponds to register files on NVIDIA GPUs. By keeping inputs and partial sums in registers or hardware-level caches, latency is further minimized. Note that in this tile program, each tile allocates the same local buffers as shared memory, which might seem counterintuitive, as shared memory is generally faster but more abundant, whereas register files is limited. This is because the allocation here refers to the register files for an entire thread block. \oursys{} uses a Layout Inference Pass during compilation to derive a Layout object \code{T.Fragment}, which determines how to allocate the corresponding register files for each thread. This process will be discussed in detail in subsequent sections.
\end{itemize}

Data transfer between global memory and hardware-specific memory can be managed using \code{T.copy}. Furthermore, hardware-specific buffers can be initialized using \code{T.clear} or \code{T.fill}. For data assignments, operations can also be performed in parallel using \code{T.Parallel}, as demonstrated in ~\ref{fig:layout_inference}.

\subsection{Dataflow Centric Tile Operators}

\begin{figure}[!htbp]
    \centering
    \vspace{-0.2cm}
    \includegraphics[width=\linewidth]{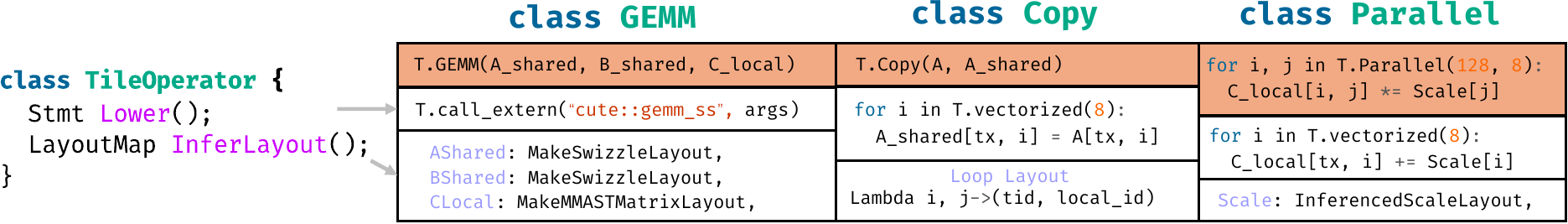}
    \vspace{-0.5cm}
    \caption{Interface of a Tile-Operator, and example instances of TileOP.}
    \vspace{-0.2cm}
    \label{fig:tile_op_examples}
\end{figure}

\oursys{} abstracts a set of Tile Operators that allow developers to focus on the dataflow logic without needing to manage the low-level implementation details of each tile operation. Figure~\ref{fig:tile_op_examples} illustrates the interface of a Tile Operator along with several representative examples, including \texttt{GEMM}, \texttt{Copy}, and \texttt{Parallel}. Each Tile Operator is required to implement two key interfaces: \texttt{Lower} and \texttt{InferLayout}. The \texttt{Lower} interface defines how the high-level Tile Operator is lowered into a lower-level IR, such as thread bindings or vectorized memory accesses. For example, \texttt{Copy} can be lowered into a loop with explicit thread binding and vectorized loads/stores. The \texttt{InferLayout} interface is responsible for determining the memory and loop layouts associated with the Tile Operator. This includes inferring buffer layouts (e.g., swizzled memory) or loop-level layouts (e.g., thread  bindings). For instance, \texttt{T.gemm} applies swizzled layouts to its shared memory inputs and uses a matrix-specific layout for writing back MMA fragments. Similarly, the parallel loop structure in \texttt{T.Parallel} can be expressed using thread-level bindings and vectorized access patterns, both of which are derived via layout inference. Section~\ref{sec:memory_layout_composition} provides a more detailed discussion of layout composition and its role in the lowering process.

Table ~\ref{tab:operators_and_primitives} lists a subset of \oursys{} operators to simplify common operations in tile-based programming. These built-in operators abstract low-level details of hardware memory access and computation, allowing developers to focus on high-level algorithm design from dataflow perspective while maintaining fine-grained control over performance-critical aspects. Each operator is designed to integrate seamlessly with the tile programming model, ensuring efficient data movement and computation across the hardware memory hierarchy. Below, we describe several key operators along with their roles in optimizing memory transfers and arithmetic computations.

\begin{itemize}
    \item \textbf{copy}: The copy op is a sugar syntax for \code{T.Parallel} with memory copy, which allows copy from and into scope fragment for registers, shared scope for static shared memory, shared.dyn for dynamic shared memory, and global for global memory.
    \item \textbf{gemm}: The built-in \code{T.gemm} operator is a highly optimized implementation for general matrix multiplication, supporting various memory access patterns (\code{ss}, \code{sr}, \code{rs}, \code{rr}), where \code{r} denotes register memory and \code{s} denotes shared memory. The operator automatically selects the optimal implementation based on the kernel configuration. For CUDA backends, \code{T.gemm} utilizes Nvidia's CUTLASS library to efficiently leverage Tensor Cores or CUDA Cores, while for AMD GPUs, it employs both composable kernels and hand-written HIP code for performance optimization. Users can also extend \code{T.gemm} by registering custom primitives in Python, making it flexible for specific use cases.
    \item \textbf{reduce}: The \code{T.reduce} operator provides a flexible and efficient reduction mechanism for aggregating data across dimensions. It supports a variety of reduction operations such as \code{sum}, \code{min}, \code{max}, and \code{product}, among others. The reduction can be performed across specified axes, enabling operations like row-wise or column-wise reductions in a matrix. \code{T.reduce} is implemented to utilize warp-level and block-level parallelism for optimal performance on both CUDA and AMD backends. Users can also customize the reduction operation by defining their own reduction kernels.
    \item \textbf{atomic}: The \code{T.atomic} operator provides atomic operations for safe updates to shared or global memory in a parallel context. Common atomic operations like \code{add}, \code{min}, and \code{max} are supported out-of-the-box. \code{T.atomic} ensures thread safety during concurrent updates, making it essential for operations like histogram updates, reductions with shared memory, and synchronization-free counters. It is designed to leverage native hardware atomic instructions on both NVIDIA and AMD GPUs, ensuring high performance while maintaining correctness in parallel executions.
\end{itemize}

\subsection{Schedule Annotations and Primitives}

While dataflow patterns form the foundation of computation organization, modern high-performance computing demands more fine-grained control over execution patterns. To address this need, \oursys{} provides a comprehensive suite of scheduling primitives that enable developers to precisely tune performance-critical aspects of their applications, as detailed in Table~\ref{tab:operators_and_primitives}:

\begin{itemize}
    \item \textbf{Pipelined}: The \code{T.Pipelined} primitive allows efficient pipelined execution of loops to improve performance by overlapping computation and memory operations. In Figure~\ref{fig:matmul_example}, the loop iterating over \code{k} (the reduction dimension) is pipelined with \code{num_stages=3}, creating a 3-stage pipeline. This pipeline allows data transfer, computation, and subsequent data preparation to overlap, effectively reducing memory bottlenecks and improving computational throughput. The detailed design for lowering the process from \code{T.Pipelined} into CUDA source code will be discussed in Section~\ref{subsection:software_pipline_jit}.

    \item \textbf{Parallel}: The \code{T.Parallel} primitive enables automatic parallelization of loops by mapping iterations to threads. In Figure~\ref{fig:layout_inference}, the operation copying data into \code{A_shared} uses \code{T.Parallel(8, 32)} to parallelize across both the \code{8} and \code{32} dimensions. It not only improves performance by leveraging hardware parallelism but also automatically maps threads to iterations and supports vectorization for further optimization.

    \item \textbf{annotate\_layout}: The \code{T.annotate_layout} primitive enables you to specify memory layout optimizations for shared or global memory using a user-defined memory layout. By default, \oursys{} adopts an optimized memory layout designed to minimize bank conflicts on both Nvidia and AMD GPUs.

    \item \textbf{use\_swizzle}: The \code{T.use_swizzle} primitive improves L2 cache locality by enabling swizzled memory accesses. improving the data reuse for rasterization. This primitive is particularly effective when processing tiled data in parallel threads blocks.

\end{itemize}

\section{Scheduling Design and Automation}
In this section, we discuss four types of schedule spaces and their automation design in TileLang besides Dataflow. Some of these are relatively independent (such as pipeline and tensorization), while others are more coupled, such as Thread Binding and Memory Layouts design. In the following sections, we will first explain the design of Memory Layout Infrastructure, followed by Thread Binding. Then, we will discuss the automation design for Tensorization, and finally share the design of Pipeline.

\subsection{Memory Layout Composition}
\label{sec:memory_layout_composition}

In \oursys{}, we support indexing into multi-dimensional arrays using a high-level interface such as \texttt{A[i, k]}. This high-level indexing is ultimately translated into a physical memory address through a series of software and hardware abstraction layers. To model this index translation process, we introduce key abstraction \textbf{Layout}, which describe how data is organized and mapped in memory. At the physical address level, a layout can be represented as a linearized address expression of the form $\sum_{i} y_i s_i$, where $y_i$ denotes the index along the $i$-th dimension, and $s_i$ is the stride that dimension contributes to the overall linear memory address. Given a layout \( L = s : d = (s_0, s_1, \ldots, s_{n-1}) : (d_0, d_1, \ldots, d_{n-1}) \), \oursys{} adopts a design inspired by TVM~\cite{tvm2018}, introducing a composable and stackable layout function abstraction built upon \textit{IterVar}. Since an \textit{IterVar} can encapsulate stride information, layout expressions can be simplified into algebraic forms over IterVars. Consequently, a layout function can be formally expressed as a mapping \( f : \mathbb{K}^n \to \mathbb{K}^m \), where \( f \) encodes the transformation from high-level indices to memory addresses.

\begin{figure}[!htbp]
    \centering
    \vspace{-0.2cm}
    \includegraphics[width=\linewidth]{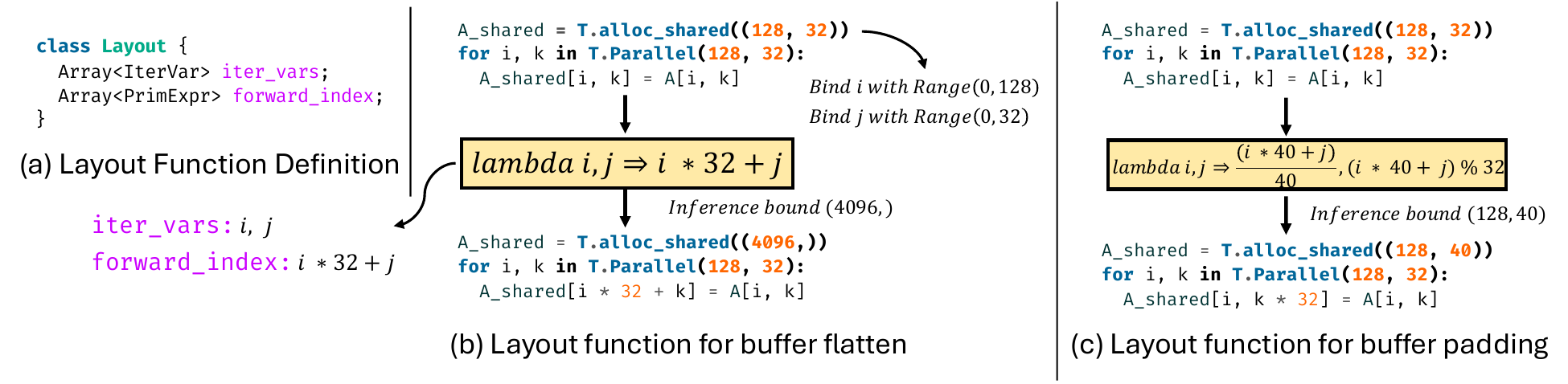}
    \vspace{-0.5cm}
    \caption{Interface and example instances of Layout Function.}
    \vspace{-0.2cm}
    \label{fig:layout_example}
\end{figure}

Figure~\ref{fig:layout_example}(a) illustrates the definition of a \texttt{Layout} in \oursys{}. Its core components include \texttt{iter\_vars}, which may optionally carry range information, and a set of \texttt{forward\_index} expressions that compute memory locations based on those iteration variables. These expressions collectively define an algebraic function \( f : \mathbb{K}^n \to \mathbb{K}^m \) . As shown in Figure~\ref{fig:layout_example}(b), this allows expressing a 2D-to-1D layout transformation. Given the shape of the buffer, \texttt{iter\_vars} are bound to specific regions, and the resulting expressions are passed to arithmetic analyzer to determine the symbolic or constant bounds. These bounds are used to infer the transformed buffer’s shape and to adjust buffer access indices accordingly.

\oursys{} also supports non-bijective layout transformations. For example, Figure~\ref{fig:layout_example}(c) demonstrates how layouts can be used to apply padding to buffer accesses. These layout transformations are composable, and \oursys{} includes several built-in layout strategies, such as layout swizzling, which is commonly employed to mitigate shared memory bank conflicts on GPUs.

In addition, \oursys{} introduces an extension of the \textbf{Layout} abstraction, referred to as \textbf{Fragment}. In contrast to standard layouts, a Fragment Layout always produces an output of the form \( f : \mathbb{K}^n \to \mathbb{K}^2 \), where the two output dimensions represent the thread's position within the register file and the index into the local register file, respectively. For instance, in Figure~\ref{fig:matmul_example}, the kernel allocates a register file \( C_{\text{local}} \) at the block level. However, since GPU register files must be partitioned among threads within a block, the Fragment Layout provides an accurate description of this partitioning scheme.

\begin{figure}[!htbp]
    \centering
    \vspace{-0.2cm}
    \includegraphics[width=\linewidth]{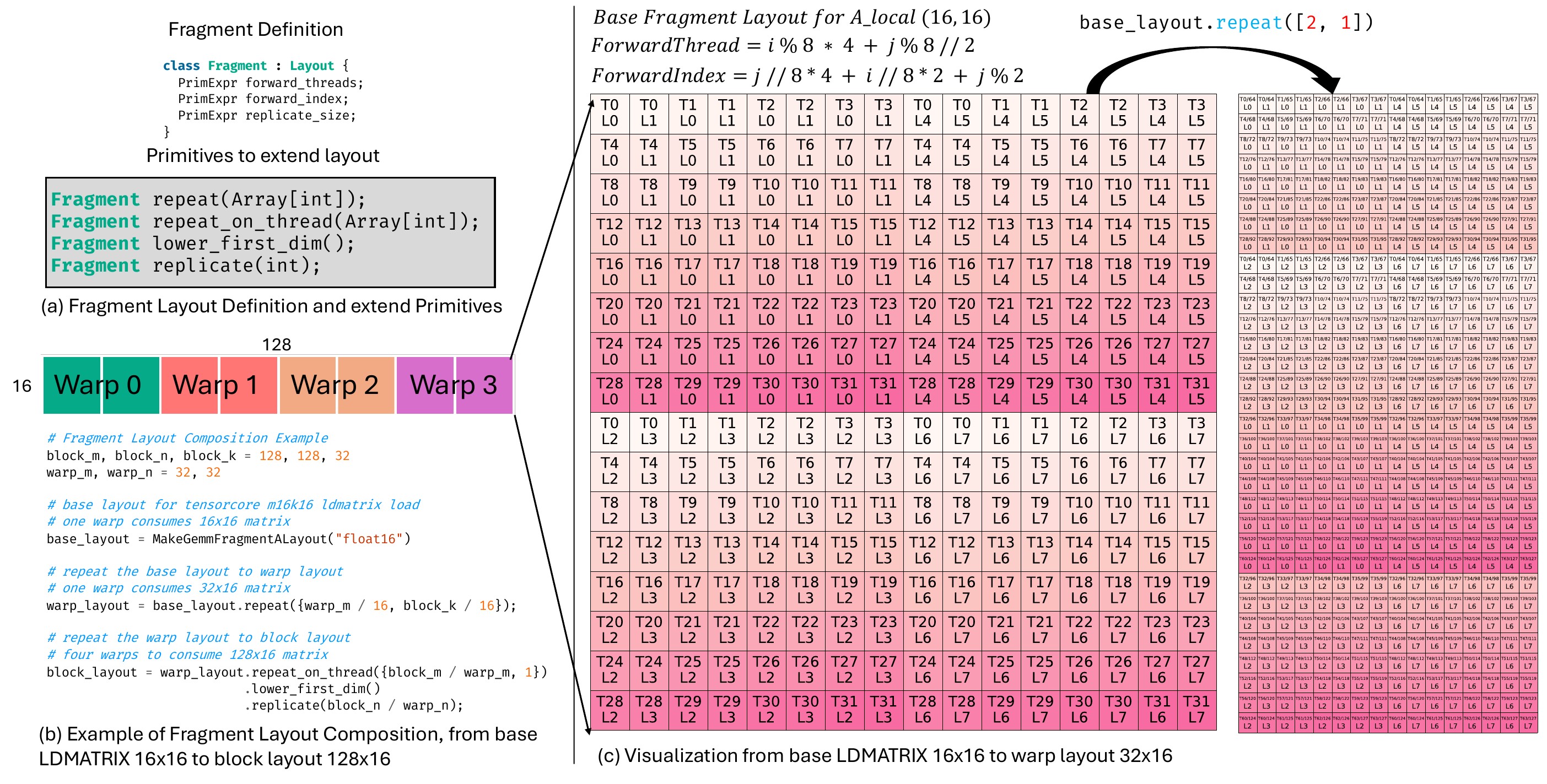}
    \vspace{-0.5cm}
    \caption{Interface and example instances of Fragment Layout.}
    \vspace{-0.2cm}
    \label{fig:fragment_example}
\end{figure}

Figure~\ref{fig:fragment_example}(a) illustrates the definition of the Fragment Layout, and \oursys{} provides four primitive operations to help users extend existing Fragment Layouts. Figure~\ref{fig:fragment_example}(b) shows an example of how these primitives are used to derive a complete block-level layout from a base layout used in the \texttt{mma\_ldmatrix} instruction for \texttt{m16k16} matrix fragments. Here, \texttt{base\_layout} denotes the layout for a single warp consuming a \texttt{m16k16} matrix. This layout is extended via the \texttt{repeat} primitive to form a \texttt{warp\_layout}, which allows a single warp to consume a \texttt{m32k16} matrix. Figure~\ref{fig:fragment_example}(c) visualizes this transformation. The \texttt{warp\_layout} is then further extended using primitives like \texttt{repeat\_on\_thread} and \texttt{replicate} to produce a \texttt{block\_layout}, which represents four warps collectively consuming a \texttt{m128k16} matrix.

\subsection{Thread Binding}

Building on the abstraction of Fragment Layouts, a key challenge that arises is how to map these layouts onto threads during execution. This leads to the \textbf{Thread Binding} problem, which involves determining how to distribute block-level register files among individual threads and how to infer appropriate fragment layouts. Moreover, it also requires identifying how loops should be correctly parallelized to match the layout constraints.

While Section~\ref{sec:memory_layout_composition} introduces Fragment Layouts to help simplify this process, determining suitable fragment layouts for all buffers remains difficult for arbitrary computational expressions. We make two key observations to guide this process. First, since multiple tile operators often share the same buffers, their respective layout and thread binding strategies are interdependent. Second, the strictness of layout and thread binding requirements varies across operators. For instance, on GPUs, the GEMM operator (which leverages Tensor Cores) imposes stringent constraints on both layout and thread binding, whereas element-wise operators typically allow more flexibility.

Based on these observations, we propose an inference scheme based on Layout and Fragment objects to optimize buffer layouts and thread bindings. To systematically manage buffer layouts, we maintain a LayoutMap that records the layout information for all buffers. We define a hierarchical priority system for tile operator layouts, where higher priority levels indicate stricter layout requirements and greater performance impact. \oursys{} processes layout inference in a top-down manner, sequentially inferring layouts from the highest to the lowest priority levels. At each priority level, \oursys{} attempts to infer layouts for all undetermined buffers until no further progress can be achieved, before proceeding to the next lower priority level.

\begin{figure}[htbp]
    \centering  
    \vspace{-0.2cm}
    \includegraphics[width=0.8\linewidth]{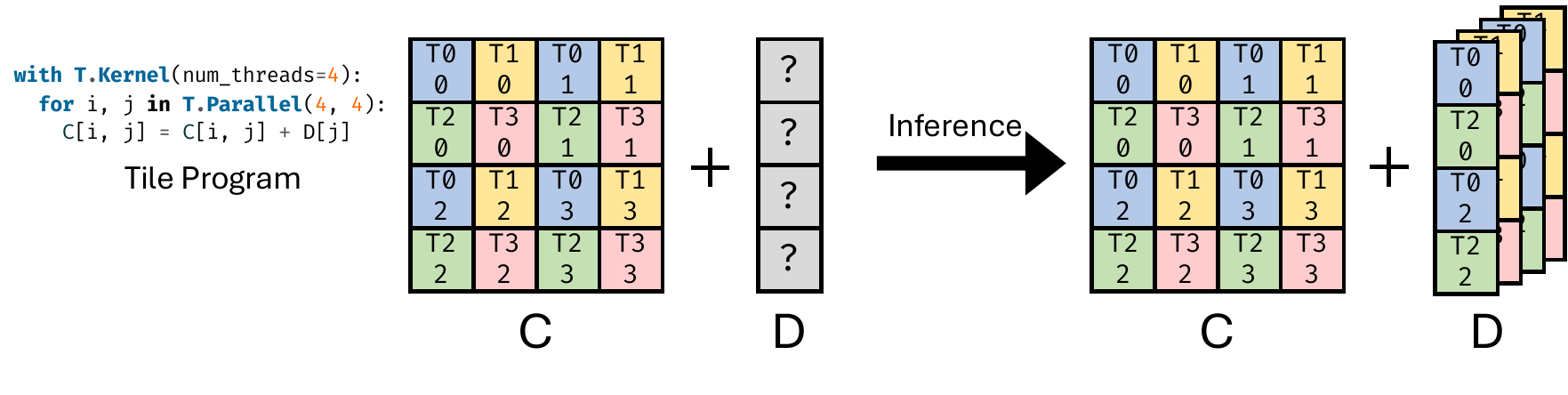}  
    \vspace{-0.5cm}
    \caption{An example of thread binding inference for Fragments.}
    \vspace{-0.2cm}
    \label{fig:fragment}
\end{figure}

As illustrated in Figure~\ref{fig:fragment}, consider a scenario where matrix C represents the result of a GEMM operation, corresponding to a Fragment object, which requires the addition of bias D post-GEMM computation. Given that GEMM holds the highest priority during the inference process, its thread binding configuration is predetermined, whereas the thread binding strategy for D remains to be determined. The output matrix C has dimensions of 4×4, distributed across 8 threads with each thread responsible for 2 elements. Consequently, the layout of the bias buffer D must be aligned with this configuration. Since each row of tensor C is processed by 2 threads, both threads require access to identical elements from D for the addition operation. Thus, D must be replicated to ensure that each thread can access the corresponding elements. The layout of D can be inferred using the same methodology.

\begin{figure}[htbp]
    \centering  
    \vspace{-5mm}
    \includegraphics[width=\linewidth]{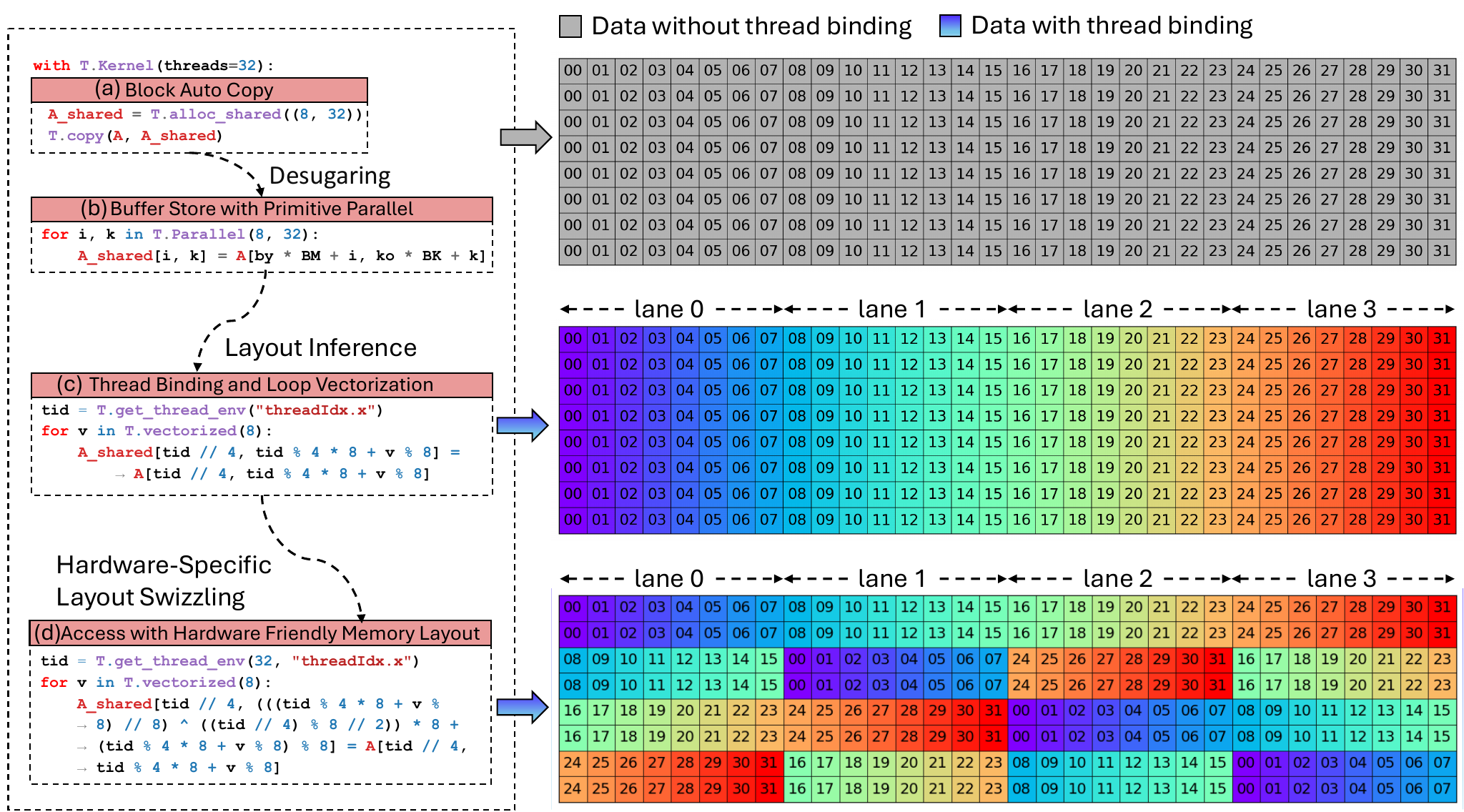}  
    \vspace{-5mm}
    \caption{Multi-Stage Automatic Thread Binding Inference for Efficient Parallel Memory Access.}  
    \label{fig:layout_inference}
\end{figure}

Figure \ref{fig:layout_inference} illustrates an example of the thread binding inference process. In particular, Figure \ref{fig:layout_inference}(a) presents a simple code snippet for copying data, which describes the dataflow of a subtile being transferred from global memory to shared memory. Proper thread binding and vectorized access can fully exploit the parallelism of GPUs and take advantage of high-performance memory access instructions. In Figure \ref{fig:layout_inference}(b), the \code{T.copy} operation is expanded into multiple loop axes. After applying the Layout Inference Pass, as shown in Figure \ref{fig:layout_inference}(c), the program undergoes automatic vectorization and parallelization. Finally, at the stage depicted in Figure \ref{fig:layout_inference}(d), Layout Swizzling is applied.

\subsection{Leveraging High-Performance Hardware Instructions}

Modern hardware architectures often support multiple instruction pathways for implementing the same computational operation. On NVIDIA GPUs, for instance, an 8-bit multiply-accumulate operation can be realized through several types of instructions. The \texttt{IMAD} instruction performs a scalar fused multiply-add operation, computing $\textstyle d = a \cdot b + c$, where all operands are internally promoted to 32-bit integers for computation. The \texttt{DP4A} instruction enables a vectorized dot-product operation, evaluating $\textstyle d = \langle \mathbf{a}, \mathbf{b} \rangle + c = \sum_{i=0}^{3} a_i b_i + c$, where $\mathbf{a}$ and $\mathbf{b}$ are 8-bit integer vectors of length four, and both the bias $c$ and the output $d$ are represented in 32-bit integer precision. For higher-throughput matrix computations, the \texttt{MMA} instruction leverages Tensor Cores to perform $\textstyle \mathbf{D} = \mathbf{A} \cdot \mathbf{B} + \mathbf{C}$, where $\mathbf{A} \in \mathbb{R}^{16 \times 32}, \mathbf{B} \in \mathbb{R}^{32 \times 8}, \mathbf{C}, \mathbf{D} \in \mathbb{R}^{16 \times 8}$; in this case, $\mathbf{A}$ and $\mathbf{B}$ are 8-bit integer matrices, while $\mathbf{C}$ and the accumulated result $\mathbf{D}$ use 32-bit integer precision. On NVIDIA RTX 3090 GPUs, the throughput of these instructions is approximately 17.8 TOPS, 71.2 TOPS, and 284 TOPS, respectively. Moreover, \texttt{MMA} instructions support various shapes under the same precision setting. 

In \oursys{}, as illustrated in Figure~\ref{fig:ptx_instructions}(a) and (b), there are two approaches to invoking hardware tensor instructions. The first approach (Figure~\ref{fig:ptx_instructions}(a)) uses C++ source injection, where instructions like \texttt{dp4a} are manually wrapped using C++ templates and injected into the kernel via \texttt{T.import\_source} and \texttt{T.call\_extern}. This enables low-level control while leveraging familiar C-style syntax. The injected function is defined at the beginning of the generated code and called within the kernel. Alternatively, as shown in Figure~\ref{fig:ptx_instructions}(b), \oursys{} provides a built-in \texttt{T.ptx} primitive that allows direct emission of inline PTX instructions (e.g., \texttt{mma.m16n8k32.row.col.s32.s8.s8.s32}) inside the kernel. This provides another low-level mechanism for utilizing specialized instructions, especially for warp-level operations.

\begin{figure}[htbp]
    \centering  
    \includegraphics[width=\linewidth]{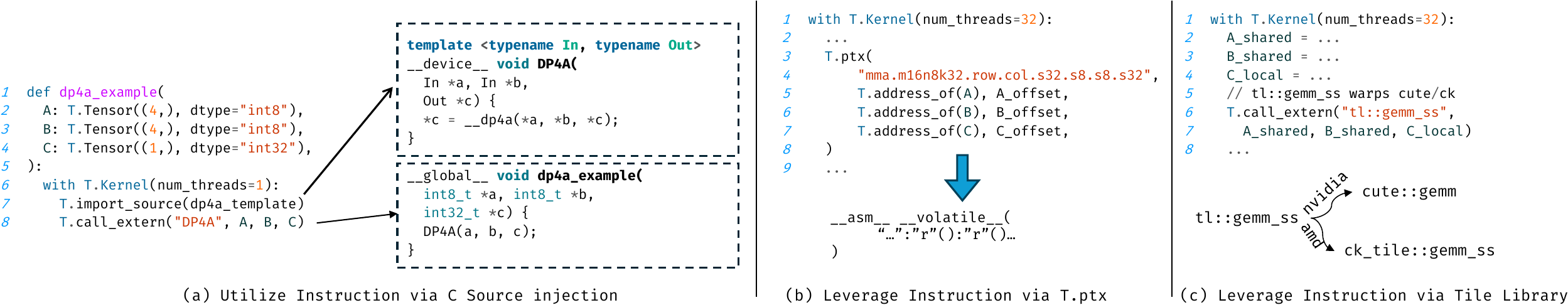}  
    \caption{Different methods of using high performance hardware instructions in \texttt{tilelang}}  
    \label{fig:ptx_instructions}
\end{figure}

However, choosing the most appropriate instruction based on input shapes and data types can be challenging. To simplify this process, \oursys{} also supports integration with Tile Libraries, as shown in Figure~\ref{fig:ptx_instructions}(c). Tile Libraries—such as NVIDIA’s \texttt{cute} or AMD’s \texttt{composable kernel (ck)}—offer high-level, standardized tile-based APIs (e.g., \texttt{tl::gemm\_ss}) for operations like GEMM. These libraries abstract away hardware-specific details and allow the underlying implementation to automatically select the most efficient instruction for a given input configuration. In \oursys{}, developers can invoke these libraries using \texttt{T.call\_extern} in a straightforward and consistent way.

In summary, \oursys{} provides two complementary methods for leveraging high-performance instructions. The first leverages Tile Libraries, which simplify integration and benefit from vendor-optimized performance. However, the high-level abstraction may limit low-level control. For example, the \texttt{cute::gemm\_ss} interface performs GEMM operations on shared memory inputs, but the data flow from shared memory to registers is internally managed by the \texttt{cute} templates. This makes it impossible to externally annotate or override internal layouts, thus reducing flexibility. Furthermore, due to heavy use of templates, compilation can become significantly slower. Analysis using the NVCC 12.8 trace tool shows that template expansion accounts for approximately 90\% of compilation time for CUDA code generated by \texttt{tilelang}.

\begin{figure}[htbp]
    \centering 
    \includegraphics[width=0.5\linewidth]{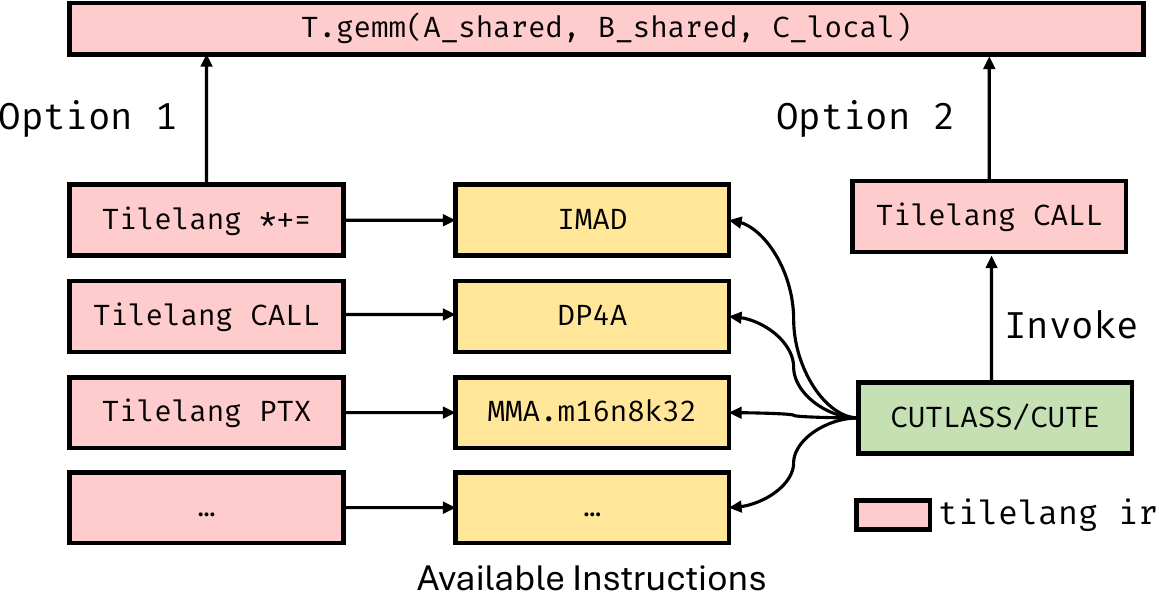}  
    \caption{Different methods of using \texttt{DP4A} and \texttt{mma} in \texttt{tilelang}}  
    \label{fig:ptx_instructions}
\end{figure}

In contrast, \oursys{} allows direct implementation of instructions via \texttt{T.gemm} using \texttt{tilelang} itself. This avoids layout annotation limitations and reduces compilation time. However, it requires users to implement a complete instruction set within \texttt{tilelang} for each target hardware instruction. Currently, \oursys{} supports both approaches, defaulting to the Tile Library-based method to facilitate rapid support for new hardware instructions.

\subsection{Software Defined Pipeline}
\label{subsection:software_pipline_jit}

\oursys{} employs an automated software pipeline inference mechanism to analyze dependencies among computational blocks (e.g., Copy and GEMM in this case) and to generate a structured pipeline schedule that maximizes parallelism while preserving correct execution order. In particular, the mechanism interleaves Copy tasks with other compute-intensive operations to reduce idle time, and when opportunities for asynchronous processing are detected, it automatically maps these tasks onto available hardware resources for concurrent execution. Consequently, \oursys{} can only expose a single \texttt{num\_stages} interface to users, significantly simplifying the process. However, we also allow users to explicitly provide information about the order and stages if needed.

\begin{figure}[!htbp]
    \centering
    \includegraphics[width=\linewidth]{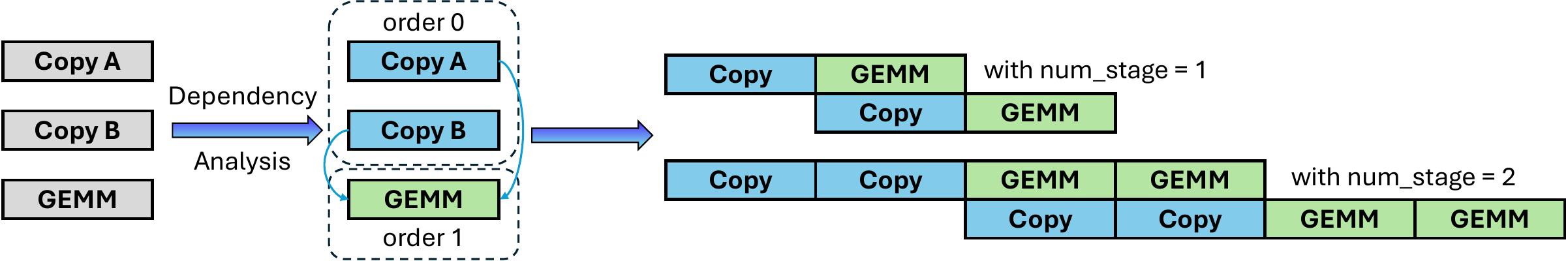}
    \caption{Software pipeline scheduling in \oursys{}. This illustration demonstrates how \oursys{} interleaves Copy and GEMM.}
    \label{fig:matmul_example}
\end{figure}

For the Ampere architecture, \oursys{} provides support for asynchronous memory copy operations using \texttt{cp.async}. The \texttt{cp.async} instruction facilitates fast data movement between global memory and shared memory, enabling overlapping of memory transfers with computation to improve performance. TileLang incorporates this capability by analyzing loop structures and automatically inserting \texttt{cp.async} instructions for eligible memory transfers. Additionally, TileLang ensures proper usage of \texttt{cp.async.commit} and \texttt{cp.async.wait} instructions to handle synchronization, guaranteeing data correctness. This optimization is particularly effective as it alleviates the pressure on register files and enables more efficient utilization of the hardware bandwidth.

In the Hopper architecture, two new features have been introduced. First, a new TMA unit is introduced as a dedicated hardware unit responsible for data copy between global memory and shared memory. Second, the PTX instruction set introduces a new wgmma instruction, which enables the execution of matrix multiplication (MMA) operations by a warpgroup (composed of four warps) to improve TensorCore utilization. Furthermore, the \texttt{wgmma.mma\_async} instruction is asynchronous. In addition, kernel optimization for the Hopper architecture commonly employs warp specialization, wherein threads are divided into producers and consumers. The producer threads use TMA to move data, while the consumer threads are responsible for the computation.

In \oursys{}, we automatically perform warp specialization optimization during the lowering process. Specifically, TileLang analyzes the buffer usage of all statements and determines their roles (producers or consumers). Based on this analysis, producers and consumers are divided into different execution paths according to threadIdx. To ensure computational correctness, TileLang leverages Live Variable Analysis to determine the appropriate synchronization points and inserts memory barriers (mbarriers) accordingly.

Asynchronous copy instructions and DMA support are also provided in the AMD CDNA architecture, which \oursys{} leverages through HIP-wrapped Copy primitives to support. Specifically, \oursys{} utilizes instructions such as \texttt{s\_waitcnt lgkmcnt} and \texttt{buffer\_load\_dword lds} to efficiently manage memory transfers. This integration enables the system to fully utilize the hardware’s capabilities for overlapping data movement with computation, further improving pipeline performance and reducing idle time.

\section{Numerical Experiments}

In this section, we evaluated the performance of \oursys{} through a series of comprehensive numerical experiments across diverse hardware platforms and workloads. Our goal is to demonstrate the effectiveness, generality, and scalability of \oursys{} in optimizing key operator kernels that form the backbone of modern machine learning workloads. By benchmarking against state-of-the-art solutions, we aim to highlight both the versatility of \oursys{} in handling mixed-precision computations and its ability to deliver significant performance gains across multiple GPU architectures.

\subsection{Experimental Setup}

\paragraph{Hardware platforms.} We evaluate \oursys{} on both NVIDIA and AMD GPUs, as they are among the most widely used accelerators. Our experiments use three cutting-edge GPUs: the NVIDIA H100 (80 GB) \cite{nvidia2023h100}, the NVIDIA A100 (80 GB) \cite{nvidia2020ampere}, and the AMD Instinct MI300X (192 GB) \cite{amd2023cdna3}. For the NVIDIA H100, we use CUDA 12.4; for the MI300X, we use ROCm 6.1.0. All platforms run under Ubuntu 20.04.

\paragraph{Operator workloads.}
We evaluate \oursys{} on a range of operator workloads that frequently appear in large-scale deep learning pipelines. On the NVIDIA H100, we focus on multi-head attention (MHA), linear attention, and general matrix multiplication (GEMM). For the NVIDIA A100, we measure performance on our dequantized GEMM kernels. Meanwhile, on the AMD Instinct MI300X, we benchmark both GEMM and MHA to capture representative use cases spanning different GPU architectures. These workloads form the foundational building blocks for many contemporary neural network models, including large language models.

\paragraph{Baselines.}

To evaluate the performance of \oursys{}, we compare it against several state-of-the-art baselines widely used in machine learning and GPU programming. These include \textbf{FlashAttention-3}, optimized for multi-head attention with CUDA instructions like \texttt{tma} and \texttt{wgmma.mma\_async}; \textbf{Triton}, an open-source framework for efficient GPU kernels that supports Nvidia and AMD GPUs but requires manual optimizations; \textbf{cuBLAS}, NVIDIA’s high-performance dense linear algebra library; AMD’s BLAS library, \textbf{rocBLAS}; \textbf{PyTorch}, featuring hand-optimized kernels like GEMM and FlashAttention-2 but not fully optimized; \textbf{BitsandBytes}, designed for supporting formats like $W_{\text{NF4}}A_{\text{FP16}}$ and provide efficient kernels; and \textbf{Marlin}, highly optimized kernels for $W_{\text{INT4}}A_{\text{FP16}}$ computations. This selection provides a comprehensive comparison across various optimization strategies and hardware compatibilities for \oursys{}.

\subsection{Experiments}

\paragraph{Flash Attention Performance.} Compared to FlashAttention-3, Triton, and PyTorch, TileLang achieves speedups of $1.36\times$, $1.41\times$, and $1.70\times$, respectively. Because FlashAttention-3 is a handcrafted approach, it cannot efficiently adapt to varying workload sizes. In particular, its fixed tile sizes cause suboptimal performance for smaller sequence lengths. For longer sequence lengths (e.g., 8k), TileLang’s performance remains close to that of FlashAttention-3. PyTorch uses a hand-optimized FlashAttention-2 kernel, which results in lower performance compared to FlashAttention-3.

\begin{figure}[!htbp]
    \centering
    \includegraphics[width=0.7\linewidth]{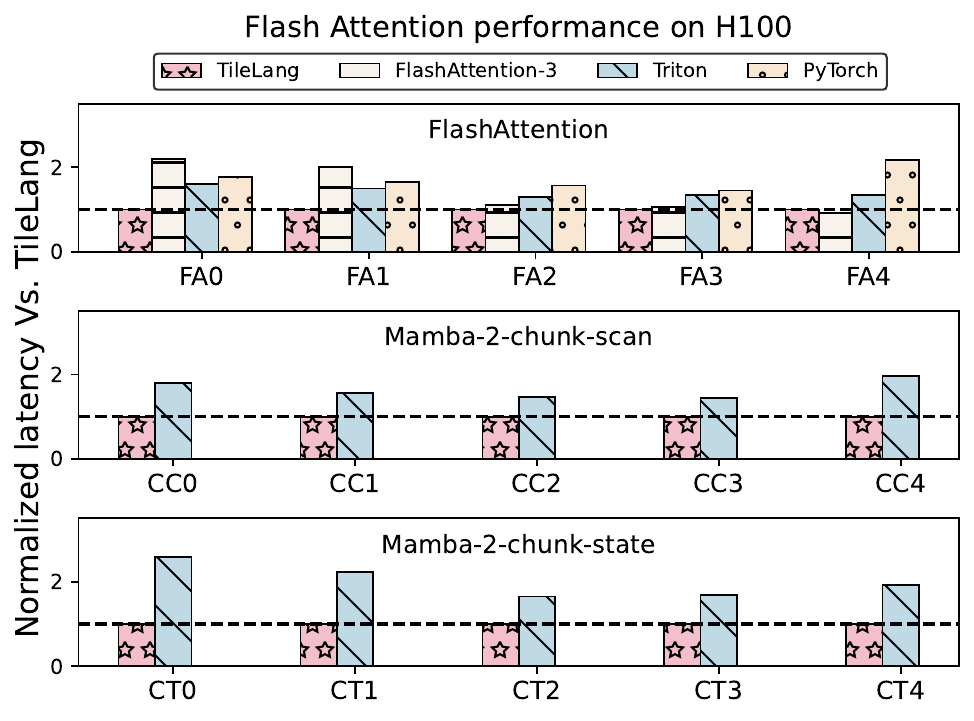}
    \vspace{-2mm}
    \caption{FlashAttention, LinearAtten Performance on Hopper Architecture.}
     \vspace{-2mm}
    \label{fig:operator_performance_mha_h100}
\end{figure}

Compared with these manually template-based implementations, TileLang can automatically utilize instructions such as \texttt{cp.async.bulk} and \texttt{wgmma.mma\_async}, and also automatically apply optimizations like warp specialization. Notably, on H100 GPUs, TileLang is capable of expressing pipeline scheduling schemes as complex as those used in FlashAttention-3.

\paragraph{Linear Attention Performance.} In our Linear Attention experiments, we use the chunk-scan and chunk-state functions from Mamba-2. Compared to Triton, TileLang achieves an average speedup of $1.77 \times$ and $2.10\times$.

\begin{figure}[!htbp]
    \centering
    \includegraphics[width=\linewidth]{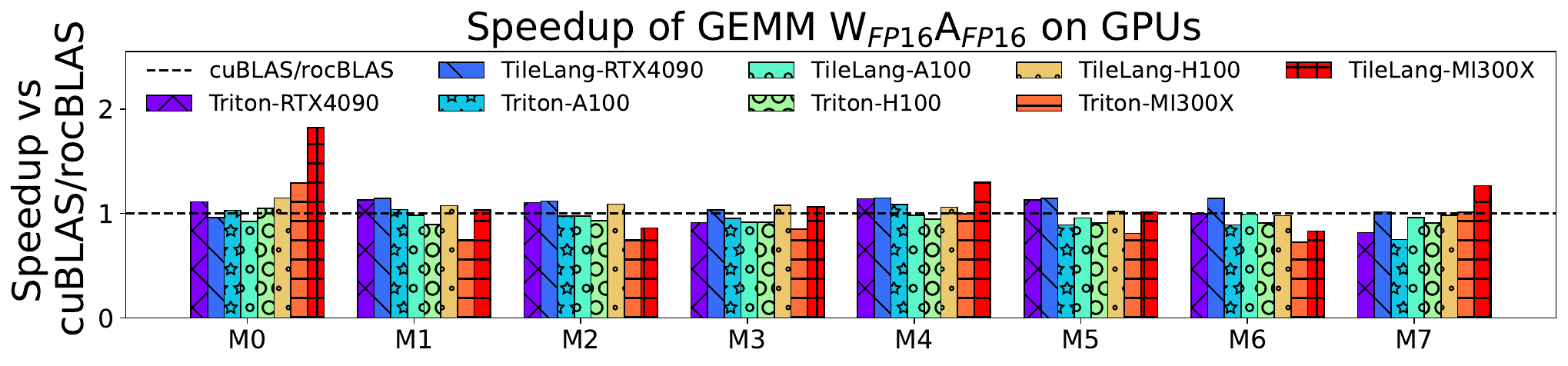}
    % \vspace{-2mm}
    \caption{GEMM performance on Nvidia and AMD GPUs.}
    \vspace{-5mm}
    \label{fig:operator_performance_matmul_nvidia}
\end{figure}

\begin{figure}[!htbp]
    \centering
    \begin{subfigure}[t]{0.48\linewidth}
        \centering
        \includegraphics[width=\linewidth]{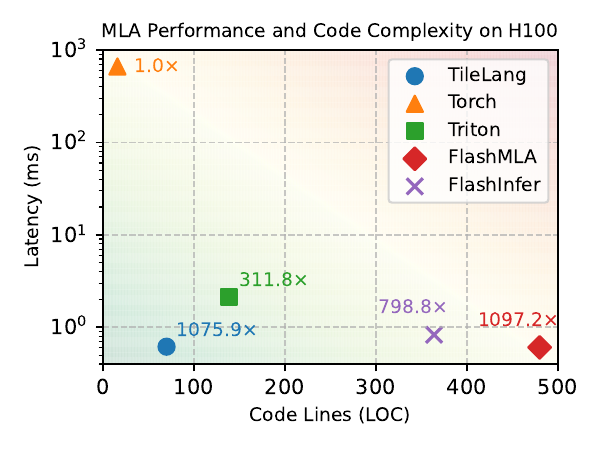}
        \caption{MLA performance and code lines on H100.}
        \label{fig:mla_h100}
    \end{subfigure}
    \hfill
    \begin{subfigure}[t]{0.48\linewidth}
        \centering
        \includegraphics[width=\linewidth]{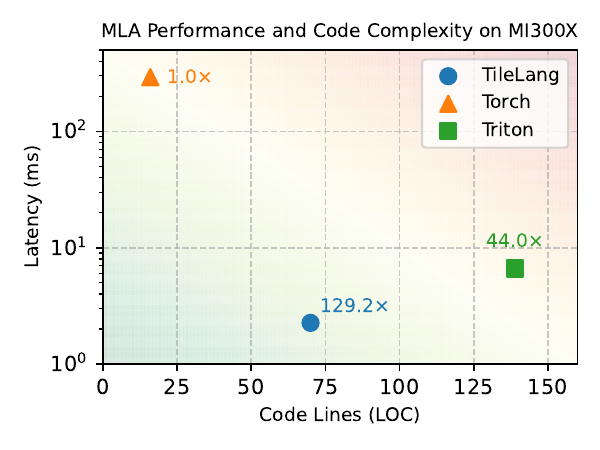}
        \caption{MLA performance and code lines on MI300X.}
        \label{fig:mla_mi300}
    \end{subfigure}
    \caption{Comparison of MLA performance and code lines on H100 and MI300X.}
    \label{fig:mla_comparison}
\end{figure}

\paragraph{Multi-Head Latent Attention Performance.}

Figure~\ref{fig:mla_comparison} illustrates the performance of MLA and the lines of code (LOC) for the corresponding kernel implementations on H100 and MI300X GPUs. On H100, \oursys{} achieves a $1075.9\times$ speedup over Torch, significantly outperforming both Triton and FlashInfer, and reaching up to 98\% of the performance of the hand-optimized FlashMLA implementation. In addition, \oursys{} requires only around 70 lines of Python code, demonstrating substantially better usability compared to other baselines. On MI300X, \oursys{} attains a $129.2\times$ speedup over Torch and surpasses Triton in both performance and code compactness. Compared to the hand-written library AITER, \oursys{} achieves 95\% of its performance. Since AITER’s kernel implementation is not open-sourced, its LOC is not included in the figure.

\paragraph{Matmul Performance.} 

Figure~\ref{fig:operator_performance_matmul_nvidia} illustrates the performance of GEMM workloads on NVIDIA and AMD GPUs, comparing \oursys{} with Triton and vendor-optimized libraries. On the RTX 4090, A100, H100, and MI300X, \oursys{} achieves speedups of $1.10\times$, $0.97\times$, $1.00\times$, and $1.04\times$ over the vendor libraries, respectively. When compared to Triton, \oursys{} delivers speedups of $1.08\times$, $1.03\times$, $1.13\times$, and $1.25\times$ on the same GPUs. For matrix multiplication, \oursys{} matches the performance of vendor-optimized libraries using a simple syntax. Additionally, by employing Layout Swizzling, \oursys{} ensures bank conflict-free execution across all tested devices.

\paragraph{Dequantize Matmul Performance.} 

\begin{figure}[!htbp]
    \centering
    \vspace{-5mm}
    \includegraphics[width=\linewidth]{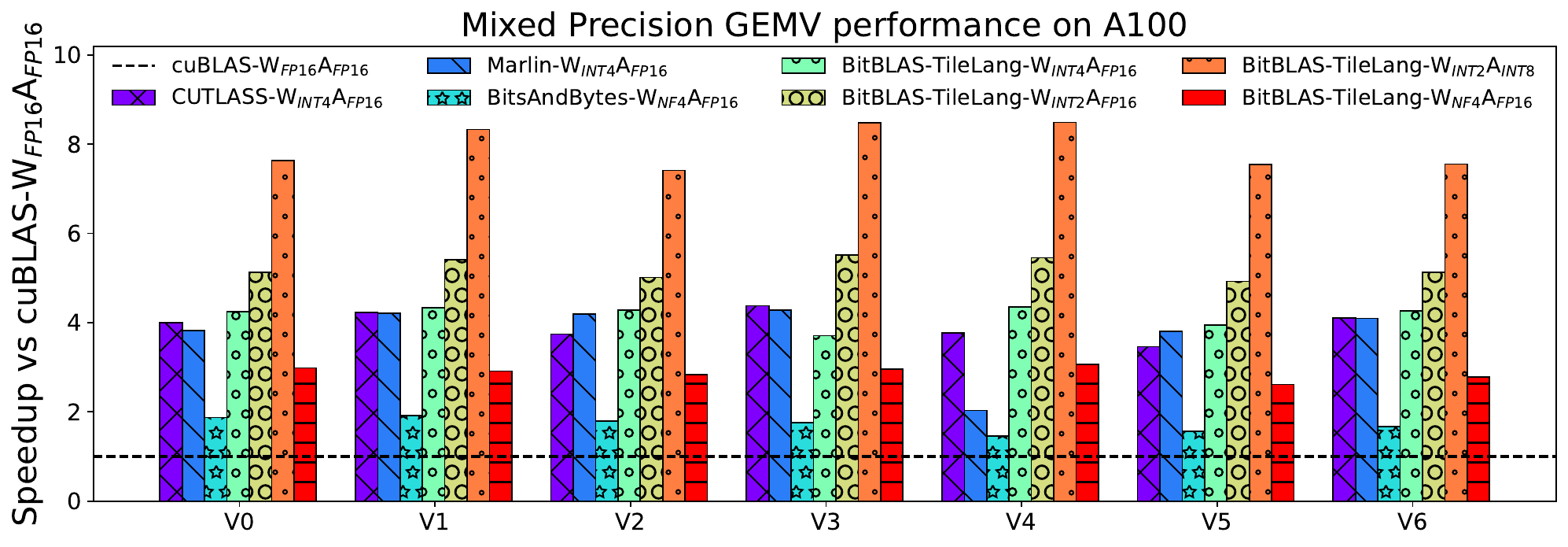}
    \vspace{-5mm}
    \caption{Dequantize Matmul Performance on A100 GPU.}
     \vspace{-1mm}
    \label{fig:operator_performance_mi300x}
\end{figure}

BitBLAS is a high-performance library for mixed-precision computations, featuring an advanced custom type system and scheduling for tensor numerical types and properties. Originally built on TensorIR, we have replaced its underlying backend with \oursys{}, enabling direct comparisons against other mixed-precision acceleration libraries. Compared to cuBLAS-$W_{\text{FP16}}A_{\text{FP16}}$, \oursys{} achieves a maximum speedup of $7.65\times$, driven by the BitBLAS-TileLang-$W_{\text{INT2}}A_{\text{INT8}}$ configuration. Additionally, for the $W_{\text{INT4}}A_{\text{FP16}}$ format, our approach delivers an average speedup of $1.04\times$ over Marlin, and for the $W_{\text{NF4}}A_{\text{FP16}}$ format, it provides an average speedup of $1.62\times$ relative to BitsandBytes. By exposing a thread-level programming interface and allowing control over data layout and pipeline configurations, \oursys{} offers developers finer-grained optimization capabilities. For example, developers can utilize PTX-based fast numerical precision conversion instructions and leverage Ladder to achieve smoother memory access within tiles. These optimizations are challenging to implement in Triton, making \oursys{} uniquely capable of delivering superior performance that Triton struggles to implement.

\section{Conclusions and Discussions}

To address the challenges of writing high-performance kernels for modern hardware accelerators, this paper introduces \oursys{}, a Python-like domain-specific language (DSL) that enables users to program at the granularity of tiles. Unlike Triton, \oursys{} allows users to explicitly declare buffers at different levels of the hardware memory hierarchy in the front-end and leverages a Layout Inference mechanism to efficiently parallelize buffer operations. This means that users only need to describe the computational logic for the buffers without worrying about how the parallelization is implemented. At the same time, \oursys{} provides the flexibility for experts to explicitly specify the exact behavior of individual threads when operating on buffers. This approach strikes a balance between ease of use and fine-grained control, offering both flexibility and performance.

Compared to ThunderKittens~\cite{thunderkittens}, \oursys{} simplifies the programming process by allowing developers to program entirely in Python while abstracting optimization details, such as pipelining, by default. For instance, in a Flash Attention implementation, \oursys{} automatically uses async copy for data movement on Ampere GPUs and lowers the pipeline to TMA on Hopper GPUs. Nevertheless, \oursys{} still provides the option for users to explicitly implement pipelining in the front-end if needed. Moreover, \oursys{} offers robust support for dynamic parameters, dynamic shapes, and other advanced features, making it particularly useful for writing kernel libraries.

We also want to discuss several promising directions exist for extending and enhancing \oursys{} in future work: First, we plan to build a self-hosting Tile Library based on \oursys{}, eliminating the current dependency on CUTLASS and manually wrapped CUDA/HIP code for built-in operators. Second, we aim to extend \oursys{} to support a range of distributed scenarios by introducing tile-level communication primitives and scheduling policies. This will allow users to implement high-performance kernels tailored to specific communication and computation resource configurations.
Additionally, we plan to investigate the design of a cost model for \oursys{}. Given the tile-based programming paradigm with explicitly exposed thread mapping details, memory access patterns and computational behaviors are clearly defined, which facilitates hardware behavior analysis and enables the development of more effective cost models. Finally, we intend to explore optimizations for dynamic shape tuning, specifically focusing on selecting the most appropriate tile configurations for programs with dynamically varying dimensions. The explicit exposure of memory hierarchies in \oursys{}’s design will further assist in supporting backends for a variety of hardware platforms, such as CPUs, NPUs, and others. We will explore a generalized design approach to extend multi-backend support, enabling \oursys{} to be seamlessly adapted to diverse hardware architectures.

Our system is open-sourced to support future development and community contributions: \url{https://github.com/tile-ai/tilelang}.

% % \label{sub:motivation_single}
% \input{code/example}

% \input{Conclusion}

%-------------------------------------------------------------------------------
\bibliographystyle{plain}
\bibliography{papers}

\appendix
\section{Operator shapes in our benchmark} \label{appendix::benchmark}

\begin{table}[h]
\centering
\resizebox{0.5\textwidth}{!}{
\begin{tabular}{c|cccccccc}
\hline
  & V0    & V1    & V2    & V3    & V4    & V5   & V6    & V7    \\ \hline
m & 1     & 1     & 1     & 1     & 1     & 1    & 1     & 1     \\
n & 16384 & 43008 & 14336 & 57344 & 14336 & 9216 & 36864 & 9216  \\
k & 16384 & 14336 & 14336 & 14336 & 57344 & 9216 & 9216  & 36864 \\
 \hline \hline
  & M0    & M1    & M2    & M3    & M4    & M5   & M6    & M7    \\ \hline
m & 4096 & 4096 & 4096 & 4096 & 8192 & 8192 & 8192 & 8192 \\
n & 1024 & 8192 & 28672 & 8192 & 1024 & 8192 & 28672 & 8192 \\
k & 8192 & 8192 & 8192 & 28672 & 8192 & 8192 & 8192 & 28672 \\
\hline
\end{tabular}
}
\caption{Matrix shapes in our benchmark.}
\label{appendix:matrix_shapes}
\end{table}

\begin{table}[h]
\centering
\resizebox{0.5\textwidth}{!}{
\begin{tabular}{c|cccccccc}
\hline
  & FA0    & FA1    & FA2    & FA3    & FA4   \\ \hline
batch & 1     & 1     & 1     & 1     & 1    \\
nheads & 32 & 32 & 32 & 32 & 32\\
seq\_len & 512 & 512 & 1024 & 1024 & 4096\\
head\_dim& 128 & 128 & 128 & 128 & 128\\
causal & true & false & true & false & true\\
\hline
\end{tabular}
}
\caption{FlashAttention shapes in our benchmark.}
\label{appendix:fmha_shapes}
\end{table}

\begin{table}[h]
\centering
\resizebox{0.5\textwidth}{!}{
\begin{tabular}{c|cccccccc}
\hline
  & CC0    & CC1    & CC2    & CC3    & CC4   & CC5\\ \hline
batch & 1     & 1     & 1     & 64     & 64     & 64 \\
nheads & 64   & 64    & 64    & 64     & 64     & 64\\
seq\_len & 1024 & 2048 & 8192 & 1024 & 2048 & 8192\\
head\_dim& 64 & 64 & 64 & 64 & 64 & 64\\
d\_state& 128 & 128 & 128 & 128 & 128 & 128\\
\hline \hline
  & CT0    & CT1    & CT2    & CT3    & CT4   & CT5\\ \hline
batch & 1     & 1     & 1     & 64     & 64     & 64 \\
nheads & 64   & 64    & 64    & 64     & 64     & 64\\
seq\_len & 1024 & 2048 & 8192 & 1024 & 2048 & 8192\\
head\_dim& 64 & 64 & 64 & 64 & 64 & 64\\
d\_state& 128 & 128 & 128 & 128 & 128 & 128\\
\hline
\end{tabular}
}
\caption{Linear Attention shapes in our benchmark.}
\label{appendix:linear_atten_shape}
\end{table}
\clearpage
\section{Kernel Implementations} \label{appendix::kernel}

\subsection{Matrix Multiplication (Matmul)}

\begin{figure}[!htbp]
\begin{minipage}{0.8\linewidth}
\begin{minted}[linenos,
               xleftmargin=\parindent,
               fontsize=\scriptsize,
               tabsize=2,
               escapeinside=||]{python}
@tilelang.jit
def Matmul(A: T.Tensor, B: T.Tensor, C: T.Tensor):
  with T.Kernel(N // block_N, M // block_M, 
    threads=threads) as (bx, by):
    A_shared = T.alloc_shared(block_M, block_K)
    B_shared = T.alloc_shared(block_K, block_N)
    C_local = T.alloc_fragment(block_M, block_N)

    T.clear(C_local)
    for k in T.Pipelined(K // block_K, num_stages=2):
        T.copy(A[by * block_M, k * block_K], A_shared)
        T.copy(B[k * block_K, bx * block_N], B_shared)
        T.gemm(A_shared, B_shared, C_local)

    T.copy(C_local, C[by * block_M, bx * block_N])
\end{minted}
\end{minipage}
\caption{Kernel Implementation of Matrix Multiplication.}
\end{figure}

\subsection{Dequantized Matrix Multiplication}

\begin{figure}[!htbp]
\begin{minipage}{0.8\linewidth}
\begin{minted}[linenos,
               xleftmargin=\parindent,
               fontsize=\scriptsize,
               tabsize=2,
               escapeinside=||]{python}
@tilelang.jit
def matmul_fp16_fp4(
    A: T.Tensor(A_shape, in_dtype),
    B: T.Tensor(B_shape, storage_dtype),
    Ct: T.Tensor((N, M), out_dtype),
):
    with T.Kernel(T.ceildiv(N, block_N), T.ceildiv(M, block_M), threads=threads) as (bx, by):
        A_shared = T.alloc_shared(A_shared_shape, in_dtype)
        B_shared = T.alloc_shared(B_shared_shape, storage_dtype)
        B_local = T.alloc_fragment(B_shared_shape, storage_dtype)
        B_dequantize_local = T.alloc_fragment(B_dequantize_shared_shape, in_dtype)
        Ct_local = T.alloc_fragment((block_N, block_M), accum_dtype)

        T.clear(Ct_local)
        for k in T.Pipelined(
            T.ceildiv(K, block_K), 
            num_stages=num_stages
        ):
            T.copy(A[by * block_M, k * block_K], A_shared)
            T.copy(B[bx * block_N, k * block_K // num_elems_per_byte], B_shared)
            T.copy(B_shared, B_local)
            for i, j in T.Parallel(block_N, block_K):
                B_dequantize_local[i, j] = _tir_packed_to_unsigned_convert("int", 8)(
                    num_bits,
                    B_local[i, j // 2],
                    j % 2,
                    dtype=in_dtype,
                )
            T.gemm(B_dequantize_local, A_shared, Ct_local, transpose_B=True)
        T.copy(Ct_local, Ct[bx * block_N, by * block_M])
\end{minted}
\end{minipage}
\caption{Implementation of Weight-Only Quantization ($W_{\text{FP4\_E2M1}}A_{\text{FP16}}$) Matmul using \oursys{}, showcasing support for mixed-precision computations via a simple form.}
\end{figure}

\clearpage

\subsection{FlashMLA Implementation}

\begin{figure}[!htbp]
\begin{minipage}{0.8\linewidth}
\begin{minted}[linenos,
               xleftmargin=\parindent,
               fontsize=\scriptsize,
               tabsize=2,
               escapeinside=||]{python}
@tilelang.jit
def flash_attn(
        Q: T.Tensor([batch, heads, dim], dtype),
        Q_pe: T.Tensor([batch, heads, pe_dim], dtype),
        KV: T.Tensor([batch, seqlen_kv, kv_head_num, dim], dtype),
        K_pe: T.Tensor([batch, seqlen_kv, kv_head_num, pe_dim], dtype),
        Output: T.Tensor([batch, heads, dim], dtype),
):
    with T.Kernel(batch, heads // min(block_H, kv_group_num), threads=256) as (bx, by):
        Q_shared = T.alloc_shared([block_H, dim], dtype)
        S_shared = T.alloc_shared([block_H, block_N], dtype)
        Q_pe_shared = T.alloc_shared([block_H, pe_dim], dtype)
        KV_shared = T.alloc_shared([block_N, dim], dtype)
        K_pe_shared = T.alloc_shared([block_N, pe_dim], dtype)
        O_shared = T.alloc_shared([block_H, dim], dtype)
        acc_s = T.alloc_fragment([block_H, block_N], accum_dtype)
        acc_o = T.alloc_fragment([block_H, dim], accum_dtype)
        scores_max = T.alloc_fragment([block_H], accum_dtype)
        scores_max_prev = T.alloc_fragment([block_H], accum_dtype)
        scores_scale = T.alloc_fragment([block_H], accum_dtype)
        scores_sum = T.alloc_fragment([block_H], accum_dtype)
        logsum = T.alloc_fragment([block_H], accum_dtype)

        cur_kv_head = by // (kv_group_num // block_H)
        T.use_swizzle(10)

        T.copy(Q[bx, by * VALID_BLOCK_H:(by + 1) * VALID_BLOCK_H, :], Q_shared)
        T.copy(Q_pe[bx, by * VALID_BLOCK_H:(by + 1) * VALID_BLOCK_H, :], Q_pe_shared)
        T.fill(acc_o, 0)
        T.fill(logsum, 0)
        T.fill(scores_max, -T.infinity(accum_dtype))

        loop_range = T.ceildiv(seqlen_kv, block_N)
        for k in T.Pipelined(loop_range, num_stages=2):
            T.copy(KV[bx, k * block_N:(k + 1) * block_N, cur_kv_head, :], KV_shared)
            T.copy(K_pe[bx, k * block_N:(k + 1) * block_N, cur_kv_head, :], K_pe_shared)
            T.clear(acc_s)
            T.gemm(
                Q_shared, KV_shared, acc_s, transpose_B=True, policy=T.GemmWarpPolicy.FullCol)
            T.gemm(
                Q_pe_shared,
                K_pe_shared,
                acc_s,
                transpose_B=True,
                policy=T.GemmWarpPolicy.FullCol)
            T.copy(scores_max, scores_max_prev)
            T.fill(scores_max, -T.infinity(accum_dtype))
            T.reduce_max(acc_s, scores_max, dim=1, clear=False)
            for i in T.Parallel(block_H):
                scores_scale[i] = T.exp2(scores_max_prev[i] * scale - scores_max[i] * scale)
            for i, j in T.Parallel(block_H, block_N):
                acc_s[i, j] = T.exp2(acc_s[i, j] * scale - scores_max[i] * scale)
            T.reduce_sum(acc_s, scores_sum, dim=1)
            T.copy(acc_s, S_shared)
            for i in T.Parallel(block_H):
                logsum[i] = logsum[i] * scores_scale[i] + scores_sum[i]
            for i, j in T.Parallel(block_H, dim):
                acc_o[i, j] *= scores_scale[i]
            T.gemm(S_shared, KV_shared, acc_o, policy=T.GemmWarpPolicy.FullCol)
        for i, j in T.Parallel(block_H, dim):
            acc_o[i, j] /= logsum[i]
        T.copy(acc_o, O_shared)
        T.copy(O_shared, Output[bx, by * VALID_BLOCK_H:(by + 1) * VALID_BLOCK_H, :])
\end{minted}
\end{minipage}
\caption{Implementation of FlashMLA with \oursys{}.}
\end{figure}

\end{document}